# Socially assistive robots' deployment in healthcare settings: a global perspective

Authors

Laura Aymerich-Franch[1]

Pompeu Fabra University

Barcelona, Spain

https://orcid.org/0000-0001-7627-5770

Iliana Ferrer

Pompeu Fabra University

Barcelona, Spain

https://orcid.org/0000-0002-7326-0415

[1]Corresponding Author:

Laura Aymerich-Franch

laura.aymerich@gmail.com

**Socially assistive robots' deployment in healthcare settings: a global perspective**

**Abstract**

One of the major areas where social robots are finding their place in society is for healthcare-related applications. Yet, very little research has mapped the deployment of socially assistive robots (SARs) in real settings. Using a documentary research method, we were able to trace back 279 experiences of SARs deployments in hospitals, elderly care centers, occupational health centers, private homes, and educational institutions worldwide from 33 different countries, and involving 52 different robot models. We retrieved, analyzed, and classified the functions that SARs develop in these experiences, the areas in which they are deployed, the principal manufacturers, and the robot models that are being adopted. The functions we identified for SARs are entertainment, companionship, telepresence, edutainment, providing general and personalized information or advice, monitoring, promotion of physical exercise and rehabilitation, testing and pre-diagnosis, delivering supplies, patient registration, giving location indications, patient simulator, protective measure enforcement, medication and well-being adherence, translating and having conversations in multiple languages, psychological therapy, patrolling, interacting with digital devices, and disinfection. Our work provides an in-depth picture of the current state of the art of SARs' deployment in real scenarios for healthcare-related applications and contributes to understanding better the role of these machines in the healthcare sector.

*Keywords:* socially assistive robots, human-robot interaction, healthcare, robot deployment

**Socially assistive robots' deployment in healthcare settings: a global perspective**

Societal challenges such as global aging or healthcare access motivate the development of socially assistive robots (SARs), which progressively enter the consumer marketplace (Breazeal, 2017; Kodate et al., 2021). SARs can be broadly defined as robots that provide assistance through social interaction (Feil-Seifer & Matarić, 2005). The social component of SARs is crucial to provide effective assistance. For a robot to be social, it needs to present specific communicative capabilities and provide social cues towards its human interaction partners to support social interaction (Hegel et al., 2009). Humans tend to anthropomorphize (i.e., attribute human characteristics) social robots (Duffy, 2003), and this creates certain expectations for the robot to behave following social behavioral patterns (Hegel et al., 2009). It could be argued that a robot becomes social in a similar manner to how a human body becomes a person: by being able to interpret and follow social rules and behaviors that allow them to interact in human environments, and by being able to recognize other humans and communicate with them. To do that, the robot needs to possess specific shapes and abilities that enable these capabilities. This does not necessarily mean a human-like shape, but one that, given its characteristics, leads humans to anthropomorphize the robot. Similarly, this does not necessarily mean that the robot needs to be able to communicate with high-level dialogue or even verbally, but that it adopts human forms of communication and interaction that are coherent with the expectations of their human interactors. Depending on their goal, it might suffice for a SAR to only be superficially socially competent or to be able to show limited social interaction capabilities. This might already significantly improve its affordances and usability (Fong et al., 2002) due to the human expectations of any robot to have social intelligence (Fincannon et al., 2004). For the purposes of the present study, the most relevant aspect to characterize a SAR is the consideration of it being specifically designed to interact with humans in a social way with the aim of providing assistance.

The unique ability of social robots to connect in a social and emotional manner with their interactors (Breazeal, 2011), makes them particularly suited for caregiving tasks (Johanson et al., 2020). SARs have also been highlighted as an opportunity to reduce the costs associated with healthcare, among others, for rehabilitation or in-home assistance-related activities (Okamura et al., 2010). For that, the commercialization of low-cost SARs with advanced features might be key to a successful and lasting implementation of this technology on a large scale (Pandey & Gelin, 2018).

In recent years, SARs have increasingly been adopted in hospitals (Van Der Putte et al., 2019), residential care facilities (McGinn et al., 2019), or private homes (Robinson et al., 2014; Torta et al., 2013), for healthcare-related purposes.

SARs for elderly care are among the most researched applications (Deutsch et al., 2019). A systematic review that examined the role of SARs for elderly care suggested that SARs have the potential to enhance elderly well-being and decrease the workload of caregivers (Kachouie et al., 2017). Another recent systematic review found that older adults would be particularly interested in employing SARs for physical assistive functions (Vandemeulebroucke et al., 2021). Within this domain, the use of SARs for Alzheimer and dementia care has received particular attention (Huschilt & Clune, 2012; Koutentakis et al., 2020). The seal robot Paro is an example of robot that has been widely explored for this purpose (Kang et al., 2020). Increasing lifelong independence is another function envisioned for SARs within this domain (Okamura et al., 2010). For instance, SARs could be used as surrogate bodies through robot avatars (Aymerich-Franch et al., 2017, 2020) to empower the elderly or physically impaired.

Another widely researched area of SARs for healthcare are cognitive-behavioral interventions and therapeutic interventions in neurodevelopmental and cognitive disorders in children (Kabacińska et al., 2020; Zhang et al., 2019). A scoping review that covered SARs as mental health interventions for children suggested that robot interventions could benefit mental health outcomes such as relief of distress and enhance positive affect (Kabacińska et al., 2020). Nao (Shamsuddin et al., 2012) or Kaspar (Wood et al., 2021) are examples of SARs that have been explored for children in the area of healthcare-related applications.

Furthermore, the Covid-19 pandemic has demonstrated that SARs can be of crucial help in the healthcare sector to develop tasks that might be dangerous for humans (Yang et al., 2020). In particular, this crisis has led to an important number of initiatives that introduced SARs in healthcare for a wide range of functions aimed at reducing contagions, such as measuring people's temperature, sanitizing areas, monitoring patients, or delivering food and medicines, among others (Aymerich-Franch, 2020; Aymerich-Franch & Ferrer, 2020b; Thomas et al., 2021).

Despite their progressive expansion, very few works have accurately examined how the deployment of SARs in human social environments is taking place. Specifically, there is an important lack of studies mapping social robot deployments for healthcare in real scenarios, as reviews are based on scientific literature, rather than real deployments.

The present work aims at palliating this gap by addressing the following research questions:

RQ1. Which robots are being deployed in real scenarios for healthcare-related purposes?

RQ2. In what areas (e.g., hospitals, homes, residential care facilities…) are social robots for healthcare being deployed?

RQ3. For what functions (e.g., entertainment, patient monitoring, telepresence...) are social robots for healthcare being deployed?

Answering these questions is important not only to monitor the deployment of this technology but also to understand the societal implications of SARs and reveal trends associated with their uses that can go beyond the purposes for which the robots were initially designed (Aymerich-Franch & Ferrer, 2020a).

Here, we examine the deployment of SARs in the area of healthcare on a global scale. By deployment, we understand that the robot has been placed in a real scenario to develop the functions for which it has been designed. For that, we conducted an exploratory study to identify in which areas, for what purposes, and with what functions are SARs being deployed. In particular, we analyzed 279 experiences of deployments involving 52 different social robot models, and taking place in 33 different countries.

## Method

In order to identify the experiences of social robots' deployments in healthcare settings, we used a documentary research method, which refers to the analysis of documents that contain information about the phenomenon we wish to study (Bailey, 1994). The documentary sources we examined included mainstream and specialized media outlets, technology blogs, robot developers' and robot suppliers' official websites and social media accounts, research papers and reports, study cases published on corporative websites, national and international robotics organizations' websites and reports, and personal communications (emails and attached materials such as flyers, images, corporative reports, or press kits, phone conversations with the actors involved in the deployments, and messages through social media). Following (Aymerich-Franch & Ferrer, 2020a, 2020b), we used a strategic non-probabilistic sample due to the large number of experiences and the impossibility to quantify all of them on a global scale.

We traced back experiences and gathered the corresponding sources between March 2020 and February 2021.

We examined more than 1.500 documentary sources and retained only the sources that described experiences that met the inclusion criteria (see Table 1) and that contained relevant information to classify the experiences.

**Table 1**

*Inclusion criteria*

| | |
|---|---|
| **Characteristics of the SARs** | |
| Social features | Only robots specifically designed to interact with humans and in human physical environments were considered |
| Embodiment | Only robots with a physical body were included |
| Appearance | Only robots with at least minimal anthropomorphic features were included. Animal-like robots (e.g., Paro) were not included |
| Communication | Only robots that exhibited at least a minimum capability to interact with humans using verbal communication were included |
| Autonomy | Only autonomous and semi-autonomous robots were included |
| Movement and navigation | Only robots with movement capacity through gesturing or autonomous navigation were included |
| **Characteristics of the experiences** | |
| Purpose | We included deployments aimed at service provision for care recipients as well as deployments aimed at education and training for students and staff in the area of health |
| Geography | Worldwide |
| Temporality | We did not include any restrictions by date in the search of the documentary sources included in the analysis |
| Size | No restriction regarding the number of robot units deployed in a specific strategy was considered |

We created a database with all the experiences that met the inclusion criteria. For each experience, we localized as many sources as possible to contrast information. In total, we included 1.264 sources in the database and up to 9 documentary sources per experience. The oldest documentary source that we kept in the database dated from 2010 and reported an experience from that year, and the newest source dated from 2021 and reported an experience from that year. Documentary sources from corporative sites and robot developers' and suppliers' sites that described the characteristics of the robots rather than specific experiences were not classified by date (441 sources). Figure 1 reports the number of sources classified by year (823 sources). As it can be observed, the largest number of sources dated from 2020.

**Figure 1**

*Number of sources identified and utilized in the final analysis classified by year*

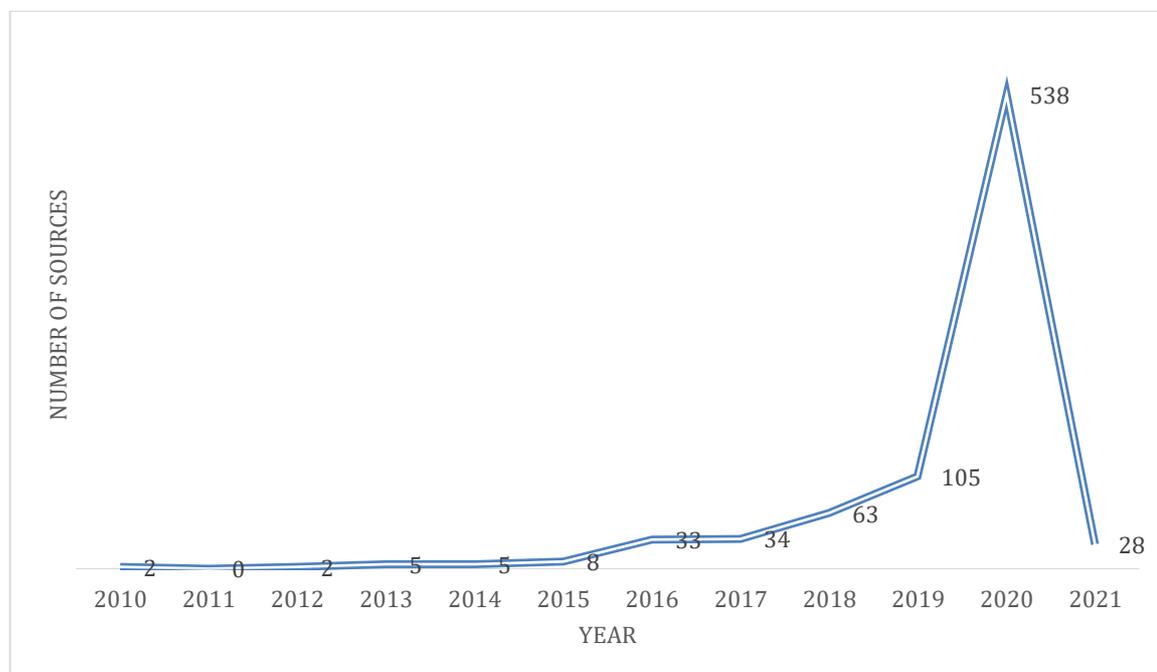

We classified each experience that was included in the database according to the following variables: social robot name, manufacturer, area and subarea, country where the experience took or is taking place, specific tasks performed by the social robot, and institution that deployed the social robot.

## Results

The final database contained 279 experiences of social robot deployments in real settings that met the inclusion criteria.

Of them, 253 experiences correspond to deployments aimed at care provision and 26 correspond to deployments aimed at education and training for medical specialties. Among the experiences aimed at care provision, we identified experiences in the following areas:

- Hospitals: this area groups experiences deployed in hospitals, medical centers, clinics, and specialized medical institutes

- Elderly care centers: this area groups experiences deployed in elderly care centers, nursing rehabilitation homes, retirement homes, aged care residences, rest homes, adult day care/support centers, homes for the elderly, and assisted living communities
- Occupational health centers: this area groups experiences deployed in occupational health centers, multidisciplinary institutions, forensic psychiatric centers, residences for people with intellectual disabilities, centers for people with special needs, institutes for autistic children and youth, and inclusive education units in primary and secondary schools
- Private homes: this area groups experiences deployed in homes of older adults, homes of children with autism or developmental disabilities, and homes of adults with different needs of assistance related to mental and physical health
- Other areas: this area groups experiences deployed in areas that do not fall into any of the previous categories. Specifically, we included experiences identified in public places, hotels, pharmacies, and schools

The experiences aimed at medical skills training were classified as a separate group. We identified experiences in the following areas:

- Educational institutions: this area groups experiences deployed in colleges, universities, university-hospitals, medical centers that train for medical specialities, academies, clinical education centers, and community colleges.

Figure 2 shows the number of experiences identified by purpose and area.

**Figure 2**

*Experiences by area (absolute number and % of total experiences)*

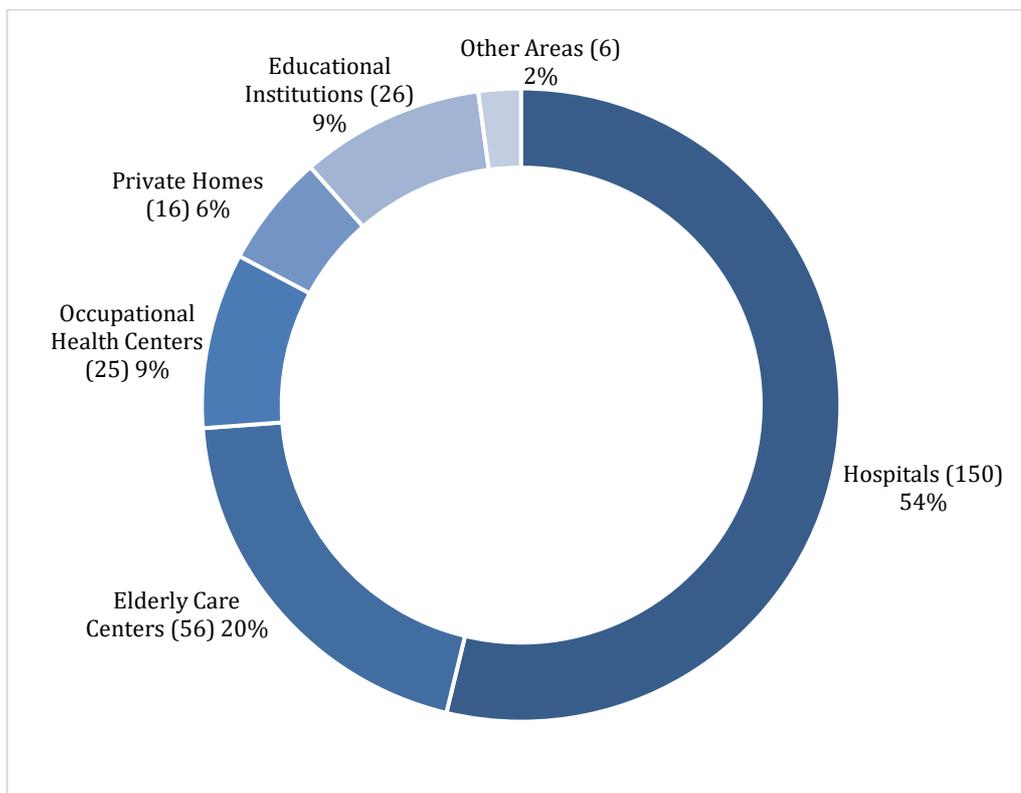

In total, 52 different models of social robots were identified. The robots with the largest number of deployments that we identified are reported in Figure 3.

**Figure 3**

*Robots (and corresponding manufacturer) with the largest number of deployments*

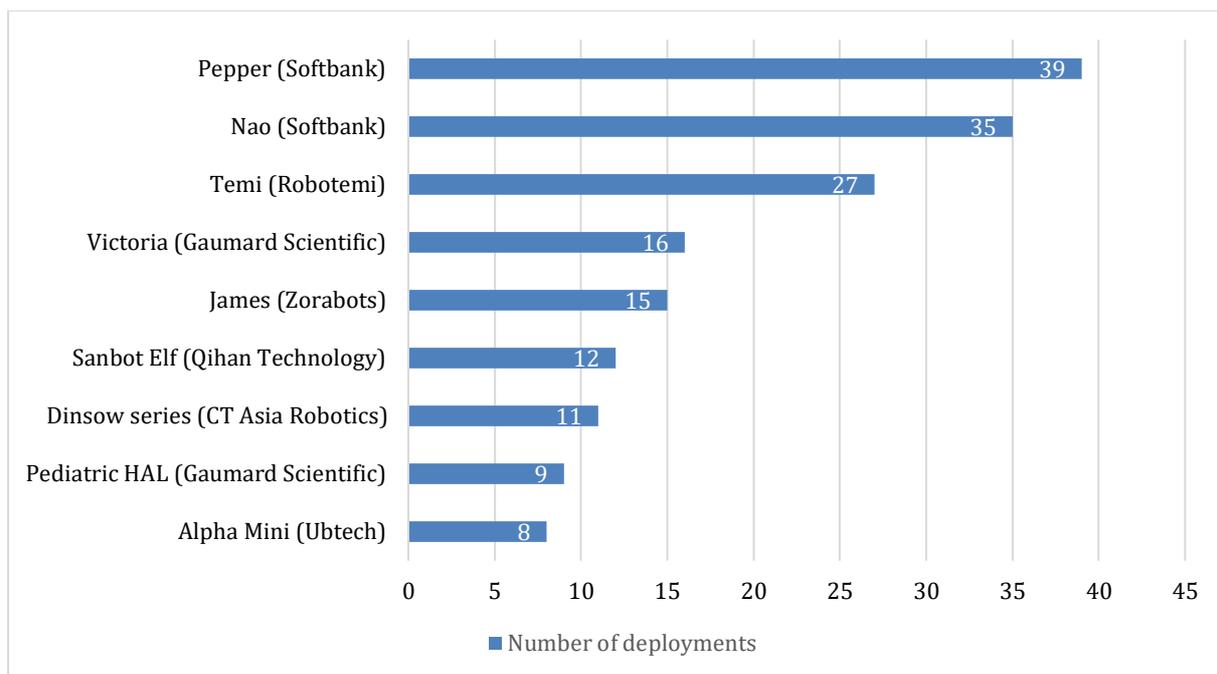

The experiences were collected across 33 countries. In particular, we report experiences from the USA (61), France (24), Belgium (23), Thailand (22), The Netherlands (22), China (19), Japan (18), Australia (13), Hong Kong (8), Spain (8), Canada (6), India (6), UK (5), Mexico (4), South Korea (4), Italy (4), Armenia (3), Germany (3), Luxemburg (3), Romania (3), Singapore (3), Finland (2), Indonesia (2), Israel (2), Sweden (2), Switzerland (2), Austria (1), Chile (1), Ethiopia (1), Hungary (1), Ireland (1), Rwanda (1), and Russia (1). Figure 4 displays a hit map of the experiences localized by country.

**Figure 4**

*Countries where the experiences were identified*

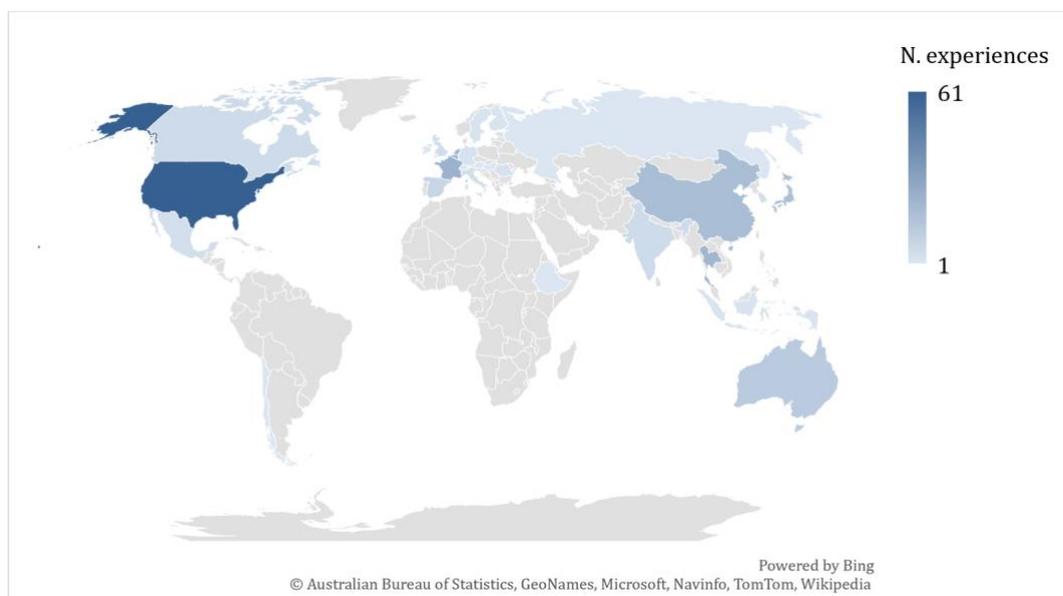

We identified a total of 20 different functions, which were extracted from the documentary sources that we analyzed. Most robots developed multiple functions. Also, in some experiences the robots were added new functions over time or initially deployed to perform certain functions and later tested for different ones. Figure 5 contains the full list of functions as well as the total number of experiences in which each function was identified.

**Figure 5**

*Functions performed by the social robots and total number of experiences in which the function was identified*

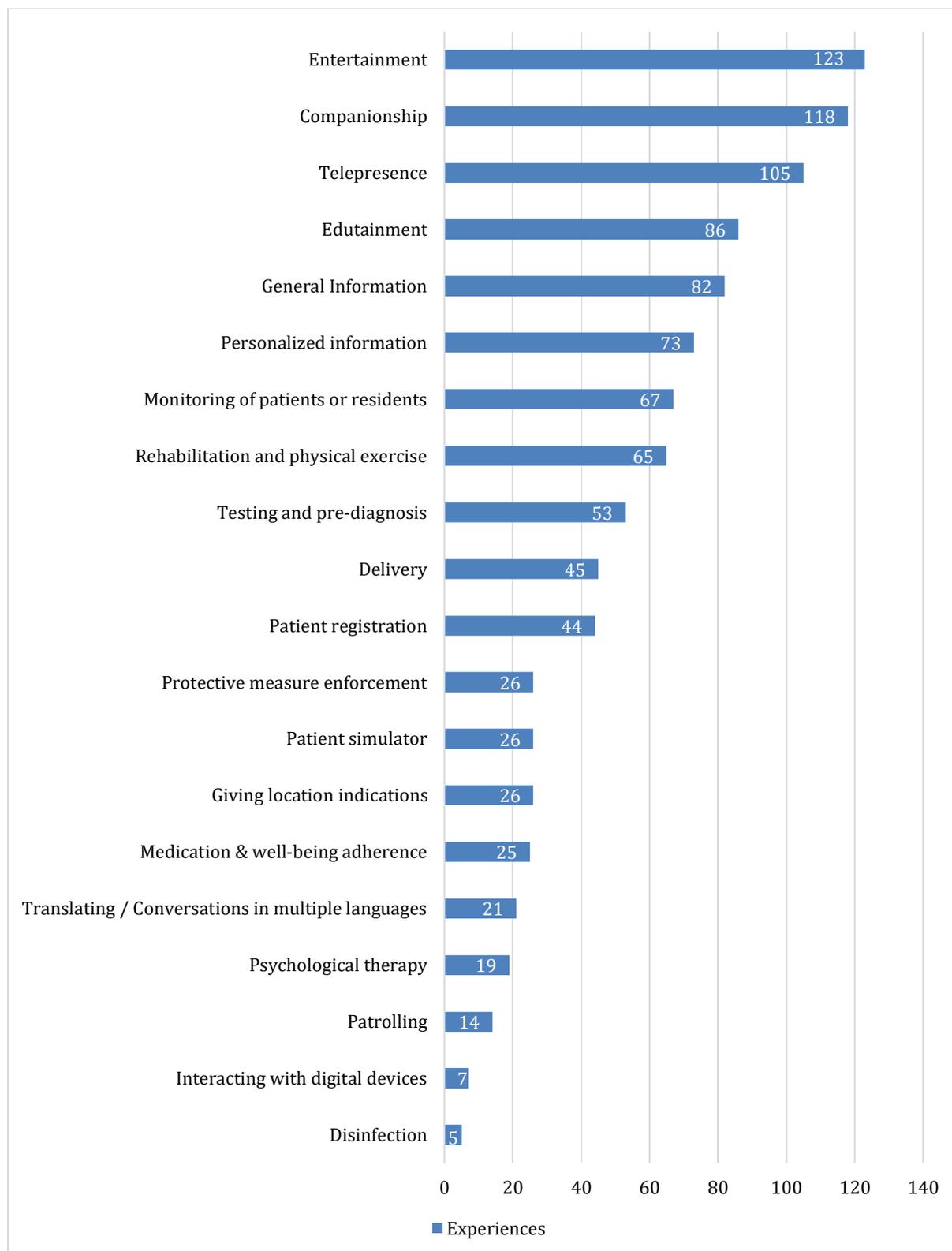

In the following sections, all the functions identified are further described and quantified by area.

**SARs deployed for care provision**

*Hospitals*

For SARs in hospitals (150 experiences), a total of 19 different functions were identified across the experiences. The most widespread function is telepresence (in 42% of the 150 experiences) for remote consultations between patients and healthcare workers. Additionally, the telepresence feature allows patients to communicate with their relatives. In some cases, the robot is used as an avatar at school for hospitalized kids to follow the lessons and be in contact with classmates and teachers through it.

Providing information is another common function. The robots can provide either general (39% of the experiences) or personalized (31%) information or advice. General information includes making announcements such as safety measures or informing schedules. Some robots are also prepared to offer personalized information or advice to patients about their medical conditions, treatments, and symptoms.

Another common function performed by SARs in hospitals is companionship (35%). Robots act as a companion offering comfort and encouraging patients' recovery by reducing loneliness and pain. Some social robots can maintain basic conversations with the patients, recognize and simulate emotions, and have physical contact with them.

Robots are also deployed for entertainment purposes (35%), motivating visitors and patients to sing, dance, or play games, to make their stay in these facilities less stressful. They also perform edutainment activities (17%). Other well-being related functions include promoting and leading physical exercise and rehabilitation (13%), improving medication adherence reminding patients to take their medicines (3%), or even in some cases providing psychological therapy (1%).

For patient monitoring (32%), the robot tracks patients' vitals and sends an alert to the medical personnel when an unusual situation is detected.

Through the testing and pre-diagnosis functions (31%), robots are able to potentially identify symptoms, or provide a preliminary diagnosis and advice before the patient is referred to the corresponding doctor. This function is principally performed by measuring vital signs and asking the patient to fill out some questionnaires to assess symptoms and it has become particularly popular during the pandemic, to track COVID-19-related symptoms (Aymerich-Franch & Ferrer, 2020b).

Some robots are deployed for delivery (25%). They assist medical staff with delivering meals, medications, internal documents, laboratory tests, medical supplies, removing dirty bedding, or providing new patients an admission pack, among others. Delivery of food and medicines is a function that has notably increased as a result of the pandemic outbreak, with the aim of reducing contagion among healthcare workers (Aymerich-Franch & Ferrer, 2020b). In fact, even some robot models that were previously identified as being deployed for waitressing functions in restaurants have now been installed in hospitals to develop this task (Aymerich-Franch & Ferrer, 2020b).

Another function is patient registration (25%), by which robots help patients registering, filling medical forms, booking appointments, preparing prescriptions, and collecting data about symptoms. These robots can also give location indications (14%) to patients and visitors or guide them to specific rooms and areas.

By translating information and having conversations in multiples languages (11%), these robots can facilitate the access of different communities to relevant health information with their capacity to translate into various languages. Multilingual robots can help medical staff to communicate with patients and visitors with language barriers, resulting in more effective interactions and better treatments' comprehension.

Other functions that were identified were patrolling (9%), which consists of keeping watch over an area by regularly navigating around it, and protective measure enforcement (15%), by which robots support surveillance duties and enforce safety measures. For instance, during the COVID-19 pandemic, they are deployed to ensure people wear masks within the premises of the hospital.

Finally, less common functions include disinfecting spaces and cleaning floors (3%), and interacting with digital devices (0.7%).

The robots localized in hospital settings among the experiences analyzed (Fig. 6) are Pepper (24) and Nao (20) – Softbank, Temi – Robotemi (20), Dinsow series – CT Asia Robotics (10), Cruzr (5) and Alpha Mini (3) – Ubtech, Greetbot – OrionStar (6), Sanbot Elf – Qihan Technology (6), James – Zorabots (3), Misty II – Misty Robotics (3), Promobot – Promobot (2), Astro – Scuola Superiore Sant' Anna (1) and Moxie – Embodied (1). Also, there were a series of robots that we identified exclusively in hospital settings, specifically, Medical Service Robot (6) and Intelligent Guidance Assistant (1) – TMI Robotics, Robin – Expper Technologies (5), Moxi – Diligent Robots (4), SOFA – The Field Robotics Institute (4), LeoBot – LionsBot International (3), WheezHope – Even Bots (3), Amy (2) and Snow (1) – Pangolin Robotics, Hospi –Panasonic (2), Sunbot-I – Siasun Robot and Automation (2), Mitra (2) and Mitri (1) – Invento Robotics, CLOi GuideBot (2) and CLOi Servebot (1) – LG Electronics, RomieBot – Roomie IT (2), XR1 – INNFOS (1), Jibo – NTT Disruption (1), AGV – OKAGV (1), Aimbot – Ubtech (1), and Sona 2.5 – ClubFirst (1).

The experiences in hospitals were identified in Thailand (22), China (19), the USA (16), Belgium (10), France (9), Japan (8), Australia (7), India (6), Canada (5), Hong Kong (5), Mexico (4), South Korea (4), Spain (4), The Netherlands (4), Italy (4), Armenia (3), Romania (3), Singapore (3), Indonesia (2), Israel (2), Switzerland (2), UK (2), Chile (1), Finland (1), Germany (1), Hungary (1), Sweden (1), and Rwanda (1).

**Figure 6**

*Example of socially assistive robots (SARs) commonly localized in hospital settings[1]*

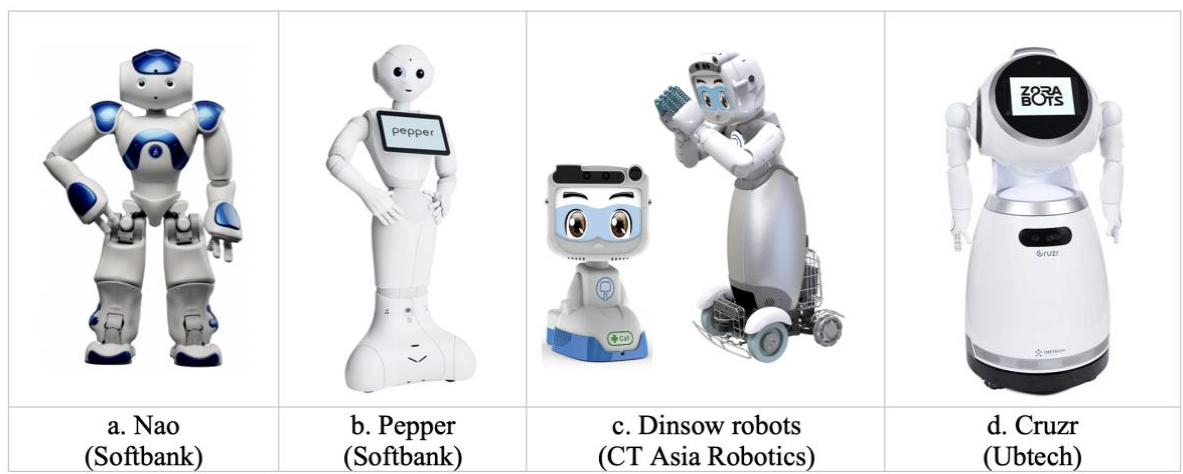

*Elderly care centers*

SARs in elderly care centers (56 experiences) perform a total of 17 different functions. In this area, the majority of tasks carried out by SARs are oriented to the residents' well-being.

The most widely available function is entertainment (identified in 84% of the 56 experiences). The robots are deployed in nursing homes for entertainment purposes like playing games, taking photos, singing, dancing, or reading news and stories.

---

[1] Credits: A) Softbank Robotics Europe, 2017, NAO Evolution , CC BY-SA 4.0
B) Softbank Robotics Europe, 2016, Pepper the Robot, CC BY-SA 4.0
C) Credited to CT Asia Robotics CO., LTD.
D) Credited to Zorabots

Companionship (66%) functions to emotionally comfort residents, and edutainment (55%), which are entertainment activities designed to be educational, are also very extended. Edutainment activities, aimed at residents and patients with Alzheimer, help stimulate cognitive abilities through memory games, dictation exercises, puzzles, quizzes, or art and language classes, among others. These activities support residents to keep mentally active, busy, and more energetic. They also work as preventive medicine and strengthen specific skills like problem-solving and critical thinking. The robots are sometimes able to recognize emotions and behavioral cues and respond accordingly, which can help patients with dementia reduce stress levels.

Another common application of SARs in elderly care centers is telepresence (57%), regarded as a solution to keep communication between the residents and their families, and strengthen the social connection.

Promoting and leading physical exercise and rehabilitation (59%) is also a widespread function. These robots can promote and demonstrate physical exercise, offer relaxation routines, meditation, rehabilitation activities, or Tai Chi and yoga programs. Some robots also count with a walking support attached to help residents get up, sit down, and walk.

Another relevant role performed by SARs is monitoring residents (16%). The cameras incorporated in the robots allow nursing staff to detect situations such as when a resident falls or also when the person presents unusual changes in their vitals.

A function that has increased in elderly care centers lately, as a result of the pandemic outbreak, is the delivery (12%) of medicines and meals, due to the need of maintaining physical distancing.

Providing general (12%) and personalized (9%) information or advice, medical and well-being adherence (12%), testing (7%), translating information and having conversations in multiples languages (7%), protective measure enforcement (7%), registering patients (2%), giving location indications (2%), patrolling (2%), and disinfecting designated areas (2%) are functions also implemented for social robots in nursing homes. Among them, the functions of patrolling, testing, and protective measure enforcement have emerged in the context of the pandemic. Through them, the robot decides for instance whether a visitor is allowed in the facility or not depending on factors such as if a potential fever is detected or whether the visitor is using a face mask (Aymerich-Franch & Ferrer, 2020b).

The robots localized in elderly care settings among the experiences analyzed (Fig. 7) are James – Zorabots (11), Nao (11) and Pepper (10) – Softbank, Sanbot Elf – Qihan Technology (6), Temi – Robotemi (6), Cutii – CareClever (5), Stevie – Akara Robotics (2), Buddy – Blue

Frog Robotics (2), AMY Service Robot – AMY Robotics (1), Astro – Scuola Superiore Sant' Anna (1), and Dinsow mini – CT Asia Robotics (1).

The experiences analyzed in elderly care centers were collected from Belgium (12), France (12), The Netherlands (9), Japan (7), the USA (4), Hong Kong (3), Australia (2), Germany (2), UK (2), Austria (1), Finland (1), and Spain (1).

**Figure 7**

*Example of socially assistive robots (SARs) commonly localized in elderly care centers[2]*

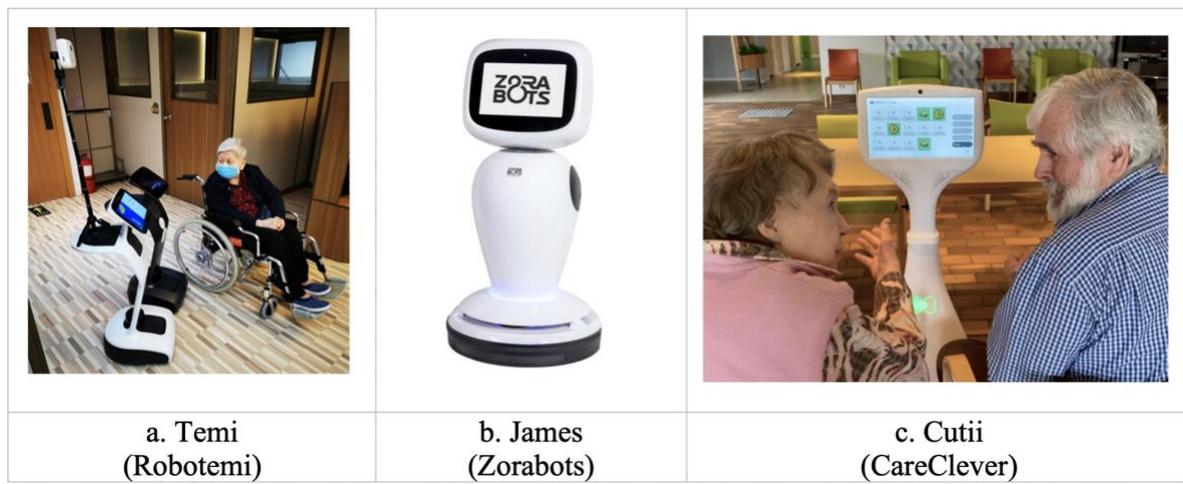

*Occupational health centers*

A total of 25 experiences were localized in occupational health centers. Social robots in these institutions are principally deployed to assist people with developmental disabilities or dementia-related patients. An initiative was also found for forensic care to help forensic patients cope with loneliness during the rehabilitation process.

In total, these robots were found to carry out a total of 13 different functions. The most relevant functions are related to support emotional well-being including providing edutainment activities (84% of the 25 experiences analyzed), psychological therapy (60%), entertainment (56%), and companionship (56%).

Robots are used to guide sessions that encourage patients' motivation to communicate, interact, self-disclose, and accept guidelines. In the case of dementia, therapists promote cognitive stimulation through SARs encouraging patients to follow edutainment tasks like reading, memory games, and other educational exercises. Also, entertainment activities are performed by the robots to help reduce stress levels and pain in those patients. For developmental disabilities, these robots are mostly deployed for children with autism spectrum disorder (ASD). Empathetic robots that act as companions can assist persons with different levels of ASD through comforting conversations and body language to train social-emotional abilities. Robots can recognize, simulate, and motivate the imitation of emotions, as well as mirror movements. Emotions or gestures can be displayed on the robot's face or screen, while movements can be shown with its arms, hands, and head. Playing games or singing in a group with the robot can also help autistic children to socialize. These robots also provide edutainment functions for children with developmental disorders such as teaching etiquette.

---

[2] Credits: A) Credited to Temi HK; B) Credited to Zorabots; C) Credited to Cutii.

Utilizing the monitoring function (8%) through the robot's camera, the therapists can also observe the patient's behavior, and use this information for improving the therapy.

For promoting physical exercise (32%), robots encourage patients to follow relaxation techniques, workout routines, as well as dancing and yoga movements. They can demonstrate how to do the exercise leading the routine and providing suggestions about how to improve.

Concerning medication and well-being adherence (8%), the robot can also remind simple routines like teeth brushing or attending therapy sessions to follow self-care activities. With the reminders, therapists intend to push patients to be more independent through settling schedule redundancy in long-term periods. Some of these robots have a small size which allows people with special needs to carry them at work, to social gatherings, or to outdoor activities, in order to assist them in their routines during the day.

Additionally, SARs also provide general (44%) and personalized (40%) information in occupational health centers giving details to patients about specific classes, programs, and schedules.

The patient registration (12%) function is used to check-in/out patients in classes or give access to personalized exercise routines that a trainer has designed previously according to the individual needs. Patients have to scan a QR code on the robot screen in order to get admission or access to the content.

Other residual functions include giving locations indications (8%), telepresence (4%), and translating and having conversations in multiple languages (4%).

The robots identified in these institutions (Fig.8) are Milo – RoboKind (5), Alpha Mini – Ubtech (4), Nao (4) and Pepper (2) – Softbank, QTrobot – LuxAI (3), iPAL – AvatarMind (3), Kaspar – University of Hertfordshire (1), Misty II – Misty Robotics (1), Moxie – Embodied (1), and Zenbo – ASUS (1).

The countries where the experiences were found are The Netherlands (8), the USA (7), Australia (4), France (2), Luxemburg (2), Spain (1) and UK (1).

**Figure 8**

*Example of socially assistive robots (SARs) in occupational health centers[3]*

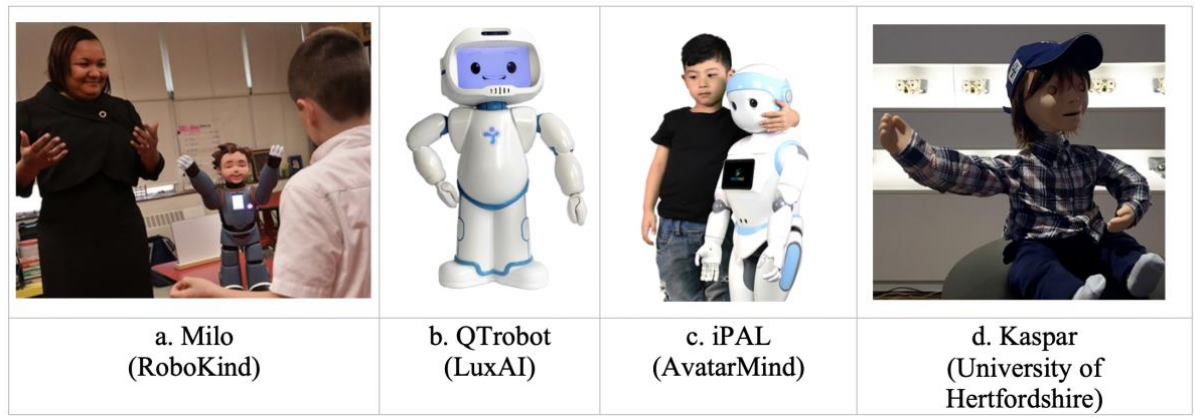

*Private homes*

We identified 16 initiatives of SARs deployed in private homes. These experiences were not counted by person/robot but by initiative (i.e., in each initiative several robots were provided to the communities participating in the projects). The organizations involved were either healthcare providers, public initiatives by city councils, associations that supported older adults to age at home, robot manufacturers and distributors that offered the robot for a monthly fee, or robot manufacturers and distributors that ceded the robots temporarily at no cost during the COVID-19 pandemic.

These robots perform 12 different functions. Among them, companionship (identified in 81% of the 16 experiences) is the most prevalent. Companion robots are essential for improving quality of life of their users. Principally, they act as empathetic conversational partners, but also provide entertainment (56%) and edutainment (56%) activities, promote and lead physical exercise and rehabilitation (25%), or even in some cases offer basic forms of psychological therapy (12%).

---

[3] Credits: A) Credited to RoboKind
B) QTrobot, 2018, QTrobot, CC BY-SA 4.0
C) Credited to AvatarMind Robot Technology
D) Matt Brown, 2017, M1llx, Science Museum - Robots - Robot for autistic children (32781592346)--modified, CC BY 2.0

Medical and well-being adherence (75%) is one of the key functionalities of these companion robots. They can be programmed to support daily routines managing the users' agenda and delivering reminders for medicines, medical appointments, exercise and meal hours, or family dates.

These robots can provide general information (6%) like the weather or news, and more personalized information (56%) as they learn about preferences and routines from the human partner.

With the telepresence (56%) feature, the users can request assistance from care-workers or relatives, make teleconsultations with doctors or nurses and get a remote diagnosis after assessing their physical condition.

SARs in this context can monitor (50%) the user and send the reports to the physician when authorized. The robots can also detect autonomously abnormal inactivity in the user or dangerous situations and transmit an alert in case help is required. Additionally, through the monitoring feature, authorized family members or nursing staff can use the video functions on the robot to navigate the house to assess the situation of the elder person in case of concern.

Finally, some of the robots offer the possibility to manage other digital devices (37%) such as smart lights or curtains and smartphones, thus robots can alert when receiving a text message, read it out loud and answer it, interact on social media, make calls, and order deliveries. Also, some are able to deliver supplies (6%).

The robots identified in these initiatives (Fig.9) are ElliQ – Intuition Robotics (5), Mabu – Catalia Health (3), Alpha Mini – Ubtech (1), Buddy – Blue Frog Robotics (1), James – Zorabots (1), Misty II – Misty Robotics (1), Moxie – Embodied (1), QTrobot – LuxAI (1), and Temi – Robotemi (1), and Stevie - Akara Robotics (1).

The countries where the experiences were identified are the USA (9), Belgium (1), Canada (1), France (1), Luxemburg (1), Spain (1), Ireland (1), and The Netherlands (1).

**Figure 9**

*Example of socially assistive robots (SARs) for older adults living alone[4]*

---

[4] Credits: A) Credited to Intuition Robotics; B) Credited to Catalia Health; C) Credited to Misty Robotics

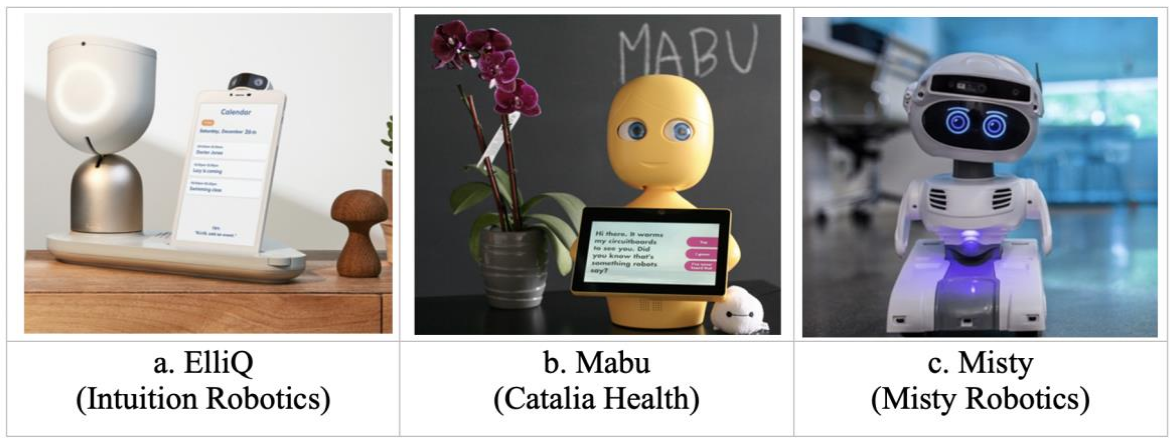

a. ElliQ (Intuition Robotics)   b. Mabu (Catalia Health)   c. Misty (Misty Robotics)

*Other areas*

Six experiences were registered in other areas, in particular, public places (2), hotels (2), a pharmacy (1), and a school (1). Despite not being deployed in a healthcare facility, these robots were included in the database because they develop functions related to public health.

In public places, these robots have been particularly promoted during the coronavirus outbreak to inform visitors about the symptoms, measure body temperature, or advise safety precautions.

Similarly, they have also been deployed in hotels for patients with mild coronavirus symptoms that need to be in isolation, where the robots can check in/out patients and provide information.

Other non-coronavirus-related initiatives placed in public buildings, pharmacies, or schools have been oriented at pre-screening diseases such as conjunctivitis, diabetes, alcoholism, and hypothyroidism, or at offering medication guidance.

The robots identified in other areas were Pepper – Softbank (3) for hotels and a pharmacy, Promobot – Promobot (1) in Times Square, Walklake – Walklake (1) in a kinder garden, and Furhat – Furhat Robotics (1) in a digital innovation building.

The countries where the experiences were found are Japan (3), the USA (1), Spain (1), and Sweden (1).

**SARs deployed for medical skills training**

*Educational institutions*

Robots as patients' simulators (26 experiences) are used for medical skills training in educational institutions. These social robots have the exclusive function of being simulators to practice medical procedures (100% of the 26 experiences analyzed).

Robots as patients' simulators are ultrarealistic androids used to train medicine, nursing paramedic, and medical specialties students free of risk in several steps of the medical practice, such as diagnosing an illness, performing treatments, assisting child deliveries, or applying resuscitation and intubation techniques. Depending on the practice for which they are intended, the androids can answer questions about their symptoms, react to stimuli, blink, cry, bleed, moan, complain of pain, track objects with their eyes, have organ sounds and pulse, simulate facial expressions and emotions (e.g., fear, anxiety), or move.

The robots localized for training purposes (Fig.10) are Victoria – Gaumard Scientific (16), specifically designed for maternal and neonatal care simulation; Pedriatic HAL – Gaumard Scientific (9), for pediatric care simulation; and Robo-C – Promobot (1), adapted to perform as a patient for an overall process of screening, diagnosis, and treatment prescription.

The experiences were collected from the USA (24), Russia (1), and Ethiopia (1).

**Figure 10**

*Example of socially assistive robots (SAR) for training in hospitals and educational institutions[5]*

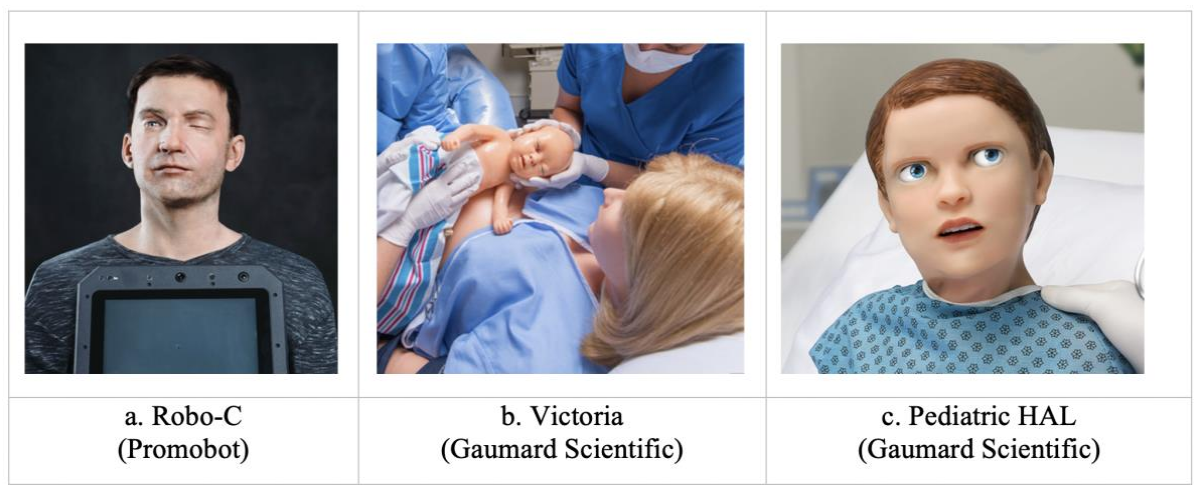

Table 2 summarizes the overall results and contains the total number of functions identified within each area, the most expanded functions by area, the robot models identified more often by area, and the countries in which more experiences were identified by area. *Other*

---

[5] Credits: A) Credited to Promobot; B) Credited to Gaumard Scientific; C) Credited to Gaumard Scientific

*areas* are not included in the table as this group comprehends miscellaneous cases from different areas and does not constitute a homogeneous category.

**Table 2**

*Results by area: Total number of functions, most expanded functions, most common robot models, and countries with more experiences.*

| Area | Total N. functions identified | Most expanded functions | Most prevalent robots | Countries with more experiences |
|---|---|---|---|---|
| *Care provision* | | | | |
| **Hospitals** | 19 | Telepresence<br>Gen. Information<br>Entertainment<br>Companionship<br>Monitoring<br>Testing / Pre-diagnosis<br>Pers. Information | Pepper<br>Nao<br>Temi | Thailand<br>China<br>the USA |
| **Elderly Care Centers** | 17 | Entertainment<br>Companionship<br>Promoting exercise<br>Telepresence<br>Edutainment | James<br>Nao<br>Pepper | Belgium<br>France<br>The Netherlands |
| **Occupational Health Centers** | 13 | Edutainment<br>Psychological therapy<br>Entertainment<br>Companionship | Milo<br>Alpha Mini<br>Nao | The Netherlands<br>the USA<br>Australia |
| **Private Homes** | 12 | Companionship<br>Adherence<br>Pers. Information<br>Entertainment<br>Edutainment<br>Telepresence | ElliQ<br>Mabu | the USA |
| *Medical skills training* | | | | |
| **Educational Institutions** | 1 | Simulators | Victoria<br>Pediatric HAL | the USA |

As it can be observed in Table 2, the total number of functions identified per area in robots used for care provision ranges from 12 in private homes to 19 in hospitals. For skills training, the robots in educational institutions are used for a single function (i.e., simulators).

The functions of companionship and entertainment are among the most expanded functions in all areas related to care provision. Similarly, telepresence and edutainment also appear among the most expanded functions in nearly all areas related to care provision.

In contrast, some functions are among the most prevalent only within a specific area. In particular, monitoring and pre-diagnosis are particularly expanded in hospitals, promoting exercise in elderly care centers, psychological therapy in occupational health centers, and medication and well-being adherence in private homes.

Regarding robot characteristics, the largest robots with movement capacity such as Pepper are more popular in hospitals and elderly care centers, smaller robots with movement capacity such as Nao are prevalent across areas, except for private homes, where the most common type are desk robots, such as ElliQ or Mabu. On the other hand, robots for skills training are uniquely designed for this purpose and not found for care provision.

Finally, concerning geography, it is noticeable that whereas experiences of SARs for care provision are present in a large number of countries, SARs for skills training are more concentrated in the USA.

## Discussion

In this work, we traced back, classified, and analyzed 279 experiences that have deployed SARs for care provision in hospitals, elderly care centers, occupational health centers, private homes, and other areas, as well as SARs deployed for medical skills training in educational institutions to determine the functions that these robots develop, the areas in which they are deployed, as well as the principal manufacturers and models that are being adopted. These experiences were collected from 33 different countries and involved 52 different models of social robots.

The most widespread functions we identified for these robots were entertainment, companionship, telepresence, edutainment, providing general and personalized information or advice, monitoring, promoting physical exercise and rehabilitation, and testing and pre-diagnosis. The results seem to indicate that one of the principal aims of SARs for healthcare is to palliate loneliness, which is suggested by the fact that entertainment and companionship are the most prevalent functions, as well as to increase connection with others, which is done through the telepresence function.

The functions of edutainment, providing general and personalized information or advice, monitoring, promoting physical exercise and rehabilitation, and testing and pre-diagnosis additionally show that SARs are progressively emerging as a solution to release medical staff and caregivers from some basic tasks and assist them with patients, older adults, and people with special needs in hospitals, nursing homes, and private homes, among others.

Rather than a human replacement, these robots are to be regarded as assistants in contexts where the human taskforce is already overloaded. The growing number of elderly populations in current societies also provides an important gap for assistance that these machines can cover when caregivers are not available or are not a feasible option (Johanson et al., 2020). Robots that serve as companions can be an important support for the older adults living alone or for children with developmental needs. Well-being related functions, such as being a conversational partner, edutaining, or practicing exercise can help users improve their mental health thanks to the motivational attitude received from the robot as well as its social presence (Robinson et al., 2014). That said, the average price of SARs is currently high and not affordable to an ample part of the population. While some initiatives have promoted SARs for private homes, a significant reduction of the cost of these robots should take place for SARs to successfully fill these roles.

The deployments of SARs in occupational health centers can help reduce healthcare costs and waiting lists. SARs are not designed to replace human therapists but can act as an effective support element in different therapies and treatments, such as in the case of autism or dementia. The robot can for instance help practice the exercises prescribed by the therapist or support in daily routines. SARs also act as an attractive device to kids with autism, helpful to keep their undivided attention. That said, SARs that are capable of providing psychological therapy are still very scarce.

Robots as patients' simulators are useful practicing tools for students in health-related careers. Robots' simulators allow students to train their skills in a low-risk scenario. These robots can be designed with different functionalities to cover a wide range of different practices.

The largest number of documentary sources that we were able to identify date from 2020. This might suggest that the COVID-19 pandemic has boosted SARs' deployment in the healthcare sector due to the need for physical distancing, which is in line with what has been suggested in other works (Aymerich-Franch, 2020; Aymerich-Franch & Ferrer, 2020b). The context of the pandemic has therefore provided a unique opportunity to test whether SARs possess a real capacity of assistance for healthcare-related purposes in real scenarios through the existing functions.

Regarding geographical distribution, the results seem to indicate that the USA, France, Belgium, Thailand, The Netherlands, China, and Japan are countries particularly open to deploy SARs for healthcare-related functions. However, these results need to be interpreted with caution as our sample is not representative. Also, language barriers as well as the media coverage and ease of access to certain experiences might have determined the experiences that we were able to identify.

Another limitation associated with the method is that the documentary sources we utilized are electronic and therefore at risk of disappearing when the websites that contain the information are restructured or eliminated. Additionally, we were not able to determine the length of most experiences we identified based on the information we obtained, but only its existence in a determinate period of time.

To conclude, we encourage future works to investigate other aspects of the experiences that we identified. In particular, it would be important to determine the effectiveness of the SARs in developing the functions for which they were deployed and to examine whether these experiences generated satisfaction among the institutions that deployed them as well as among the final users.

27Running head: DEPLOYMENT SOCIAL ROBOTS HEALTHCARE